\documentclass{article}
\usepackage{ijcai24}

\usepackage{times}
\usepackage{soul}
\usepackage{url}
\usepackage[hidelinks]{hyperref}
\usepackage[utf8]{inputenc}
\usepackage[small]{caption}
\usepackage{graphicx}
\usepackage{amsmath}
\usepackage{amsthm}
\usepackage{enumitem}
\usepackage{booktabs}
\usepackage{algorithm}
\usepackage{algorithmic}
\usepackage[switch]{lineno}
\newcommand{\hide}[1]{}
\newcommand{\vpara}[1]{\vspace{0.05in}\noindent \textbf{#1 }}
\usepackage{rotating}
\usepackage{multirow} 
\usepackage{array}

\usepackage{tikz}
\usepackage[edges]{forest}
\definecolor{hiddendraw}{RGB}{205, 44, 36}
\definecolor{hidden-blue}{RGB}{99,178,238} 
\definecolor{hidden-orange}{RGB}{243,202,120}
\definecolor{hidden-yellow}{RGB}{253,225,28}
\definecolor{hidden-green}{RGB}{77,214,12}
\definecolor{hidden-grey}{RGB}{175,171,171}

\usepackage{CJKutf8}

\title{A Survey on Neural Question Generation: Methods, Applications, and Prospects}

\hide{
\author{
    Author Name
    \affiliations
    Affiliation
    \emails
    email@example.com
}
}

\author{
Shasha Guo$^1$\thanks{~~Work was done during an internship at SMU.}\and
Lizi Liao$^2$\and
Cuiping Li$^1$\and
Tat-Seng Chua$^3$
\\
\affiliations
$^1$Renmin University of China\\
$^2$Singapore Management University\\
$^3$National University of Singapore\\
\emails
\{guoshashaxing, 
licuiping
\}@ruc.edu.cn,
lzliao@smu.edu.sg,
dcscts@nus.edu.sg
}

\begin{document}

\maketitle
\begin{abstract}
In this survey, we present a detailed examination of the advancements in Neural Question Generation (NQG), a field leveraging neural network techniques to generate relevant questions from diverse inputs like knowledge bases, texts, and images. The survey begins with an overview of NQG's background, encompassing the task's problem formulation, prevalent benchmark datasets, established evaluation metrics, and notable applications. It then methodically classifies NQG approaches into three predominant categories: \textbf{structured NQG}, which utilizes organized data sources, \textbf{unstructured NQG}, focusing on more loosely structured inputs like texts or visual content, and \textbf{hybrid NQG}, drawing on diverse input modalities. This classification is followed by an in-depth analysis of the distinct neural network models tailored for each category, discussing their inherent strengths and potential limitations. The survey culminates with a forward-looking perspective on the trajectory of NQG, identifying emergent research trends and prospective developmental paths. Accompanying this survey is a curated collection of related research papers, datasets and codes\footnote{\href{https://github.com/PersistenceForever/Neural-Question-Generation-Survey-List}{https://github.com/PersistenceForever/Neural-Question-Generation-Survey-List}}, providing an extensive reference for those delving into NQG.
\end{abstract}
\section{Introduction}
\label{sec:intro}
Question Generation (QG) represents a crucial and complex task within the domain of natural language processing (NLP). Its objective is to automatically generate questions from various sources such as knowledge bases~\cite{MHQG,AutoQGS,KQG-COT}, natural language texts~\cite{CGCQG,SGGDQ,UMQAQG}, and images~\cite{DGN,MOAG,KECVQG}. The task has garnered substantial interest in the research community, attributable to its wide-ranging applications. Notably, QG serves as a means for data augmentation, enhancing the corpus of training data for question-answering (QA) tasks, thereby refining QA models~\cite{G2S,DSM}. Additionally, it plays a vital role in intelligent tutoring systems by generating diverse questions from educational materials, aiding in evaluating and fostering student's learning~\cite{EQG,QAGE}. Furthermore, QG contributes significantly to conversational systems, enabling them to initiate more engaging and dynamic human-machine interactions~\cite{ConvMR,DialogueSystem}. In the realm of fact verification, QG is pivotal in creating training claims to augment the effectiveness of verification models~\cite{ZSFVC,LLMFC}.

The ascendance of deep neural networks~\cite{transformer,shen2018disan} has prompted a paradigm shift in QG methodologies. The field has progressively transitioned from rule-based approaches to neural network-based (NN-based) methods~\cite{KTG,UMQAQG,G2S}. Predominantly, these NN-based approaches follow the Sequence-to-Sequence (Seq2Seq) framework, utilizing various encoder-decoder architectures to refine question generation. However, a critical limitation of these models is their reliance on extensive training data, a challenge exacerbated by the typically small size of benchmark datasets in QG, leading to potential overfitting issues.

The emergence of pre-trained language models (PLMs), such as T5~\cite{T5} and BART~\cite{BART}, represents a significant advancement. These models, pre-trained on extensive corpora, possess a wealth of semantic knowledge, which significantly enhances performance in various NLP tasks upon fine-tuning. Hence, PLMs effectively address the challenge faced by previous NN-based models in QG, obviating the need for training models from scratch. This development has established the pre-training-fine-tuning framework as the dominant paradigm in QG, achieving unprecedented state-of-the-art (SOTA) results.

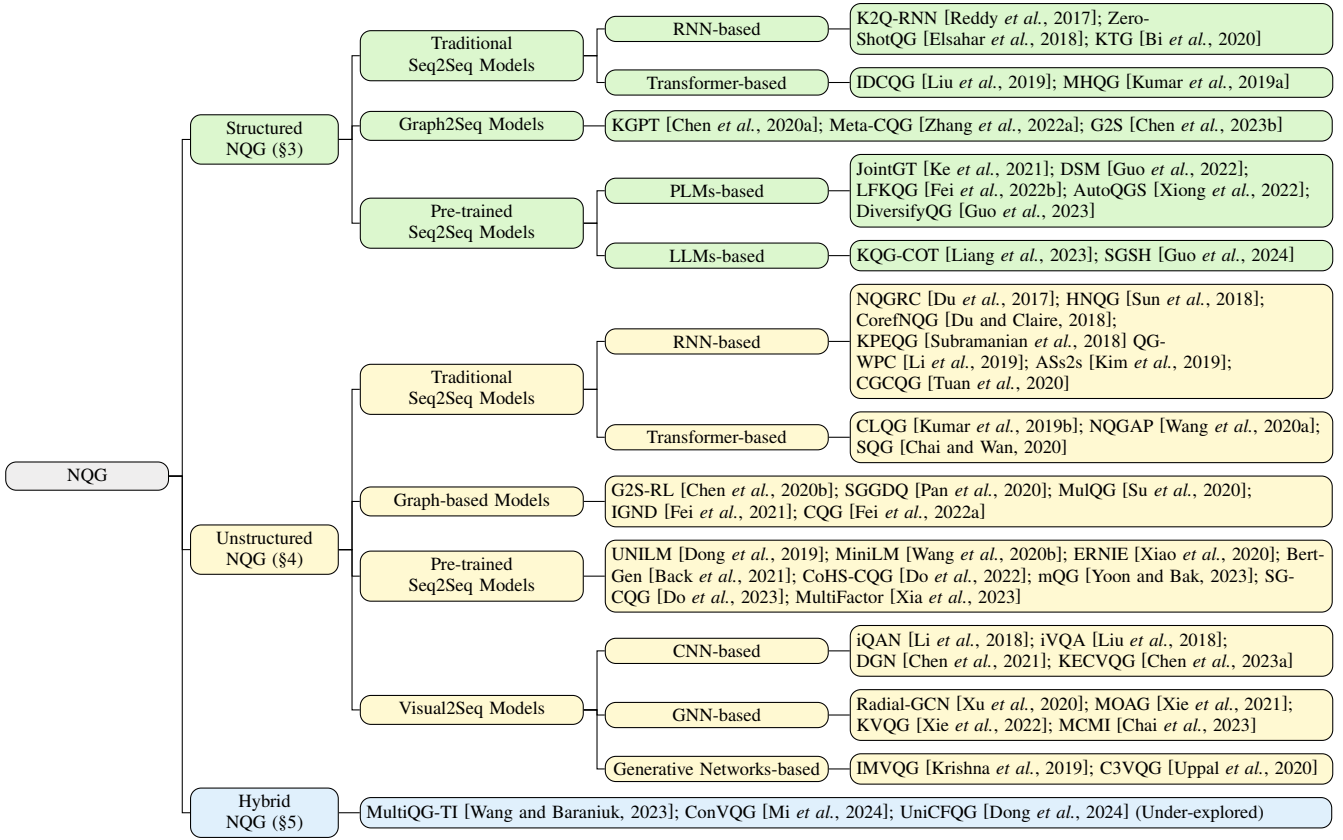
\begin{figure*}[!t]
	\scriptsize
	\begin{forest}
		for tree={
			forked edges,
			grow'=0,
			draw,
			rounded corners,
			node options={align=center},
			calign=edge midpoint,
		},
	    [NQG, text width=2cm, for tree={fill=hidden-grey!20}
			[Structured NQG (\S \ref{sec:stru_qg}), text width=1.8cm, for tree={fill=hidden-green!20}
                [Traditional Seq2Seq Models, text width=2.8cm, for tree={fill=hidden-green!20}
				 [RNN-based, text width=2.8cm, for tree={fill=hidden-green!20}
                         [K2Q-RNN~\cite{K2Q-RNN}; Zero-ShotQG~\cite{Zero-ShotQG}; KTG~\cite{KTG}, 
                          text width=6.25cm, node options={align=left}
                          ]
                        ]
                        [Transformer-based, text width=2.8cm, for tree={fill=hidden-green!20}
                         [IDCQG~\cite{IDCQG}; MHQG~\cite{MHQG}, 
                         text width=6.25cm, node options={align=left}
                         ]
                         ]              
				]
			[Graph2Seq Models, text width=2.8cm, for tree={fill=hidden-green!20}
					[KGPT~\cite{KGPT}; Meta-CQG~\cite{Meta-CQG}; G2S~\cite{G2S},
					text width=9.5cm, node options={align=left}
					]
                ]
                 [Pre-trained Seq2Seq Models, text width=2.8cm, for tree={fill=hidden-green!20}
                    [PLMs-based, text width=2.8cm, for tree={fill=hidden-green!20}
                    [JointGT~\cite{JointGT}; DSM~\cite{DSM}; LFKQG~\cite{LFKQG}; AutoQGS~\cite{AutoQGS}; DiversifyQG~\cite{DiversifyQG}, 
                    text width=6.25cm, node options={align=left}
                    ]
                     ]
                      [LLMs-based, text width=2.8cm, for tree={fill=hidden-green!20}
                      [KQG-COT~\cite{KQG-COT}; SGSH~\cite{SGSH}, 
                    text width=6.25cm, node options={align=left}
                       ]
                     ]
                   ]  
			]
			[Unstructured NQG (\S \ref{sec:un_qg}), text width=1.8cm, for tree={fill=hidden-yellow!20}
				   [Traditional Seq2Seq Models, text width=2.8cm, for tree={fill=hidden-yellow!20}
                       [RNN-based, text width=2.8cm, for tree={fill=hidden-yellow!20}
                         [NQGRC~\cite{NQGRC}; HNQG~\cite{HybridNQG}; CorefNQG~\cite{CorefNQG}; 
                         KPEQG~\cite{KPEQG}
                         QGWPC~\cite{QGWPC}; ASs2s~\cite{ASs2s}; CGCQG~\cite{CGCQG}, 
                          text width=6.25cm, node options={align=left}
                          ]
                        ]
                        [Transformer-based, text width=2.8cm, for tree={fill=hidden-yellow!20}
                         [CLQG~\cite{CLQG}; NQGAP~\cite{NQGAP}; SQG~\cite{SQG}, 
                         text width=6.25cm, node options={align=left}
                         ]
                         ]
                      ]                                     
				[Graph-based Models, text width=2.8cm, for tree={fill=hidden-yellow!20}
                     [G2S-RL~\cite{G2S-RL}; SGGDQ~\cite{SGGDQ}; MulQG~\cite{MulQG}; IGND~\cite{IGND}; CQG~\cite{CQG}, 
                     text width=9.5cm, node options={align=left}
                     ]                                          
                    ]
                    [Pre-trained Seq2Seq Models, text width=2.8cm, for tree={fill=hidden-yellow!20}                 
                    [UNILM~\cite{UNILM}; MiniLM~\cite{MiniLM}; ERNIE~\cite{ERNIE-GEN}; BertGen~\cite{BertGen}; CoHS-CQG~\cite{CoHS-CQG}; mQG~\cite{mQG}; SG-CQG~\cite{SG-CQG}; MultiFactor~\cite{MultiFactor}, 
                    text width=9.5cm, node options={align=left}
                    ]                                           
                   ]  
                    [Visual2Seq Models, text width=2.8cm, for tree={fill=hidden-yellow!20}
                       [CNN-based, text width=2.8cm, for tree={fill=hidden-yellow!20}
                       [iQAN~\cite{iQAN}; iVQA~\cite{iVQA}; DGN~\cite{DGN}; KECVQG~\cite{KECVQG}, 
                      text width=6.25cm, node options={align=left}
                       ]
                       ]
                     [GNN-based, text width=2.8cm, for tree={fill=hidden-yellow!20}
                      [Radial-GCN~\cite{Radial-GCN}; MOAG~\cite{MOAG}; KVQG~\cite{KVQG}; MCMI~\cite{MCMI},
                      text width=6.25cm, node options={align=left}
                      ]
                     ]
                    [Generative Networks-based, text width=2.8cm, for tree={fill=hidden-yellow!20}
                    [IMVQG~\cite{IMVQG}; C3VQG~\cite{C3VQG}, 
                    text width=6.25cm, node options={align=left}
                    ]
                   ]                   
                  ]
			]
                [Hybrid NQG (\S \ref{sec:hybrid_nqg}), text width=1.8cm, for tree={fill=hidden-blue!20}
					[MultiQG-TI~\cite{wang2023multiqg}; ConVQG~\cite{ConVQG}; UniCFQG~\cite{Unify_FQG} (Under-explored),
					text width=12.75cm, node options={align=left}
					]
			]   
	]
	\end{forest}
	\caption{The taxonomy of NQG. We classify NQG into three types based on input modalities: Structured NQG, which deals with structured data; Unstructured NQG, which handles unstructured data; and Hybrid NQG, which integrates both structured and unstructured data.
 }
    \label{fig:taxonomy}
\end{figure*}

With the continuous scaling of PLMs in terms of parameter size and training corpus volume, the field has witnessed the evolution of large language models (LLMs) such as ChatGPT~\footnote{\href{https://openai.com/blog/chatgpt}{https://openai.com/blog/chatgpt}} and Llama2~\footnote{\href{https://ai.meta.com/llama/}{https://ai.meta.com/llama/}}. 
These models surpass PLMs in semantic richness, offering remarkable improvements across a wide array of NLP tasks~\cite{codeGeneration,Text-to-SQL}. Consequently, the research focus has shifted towards leveraging LLMs for QG, aiming to capitalize on their advanced semantic understanding~\cite{KQG-COT}. 
A pivotal aspect in this context is In-Context Learning (ICL), a unique capability of LLMs, which can be effectively harnessed through well-designed prompts to generate the desired questions.

Existing surveys on QG primarily concentrate on traditional Seq2Seq models and on models that generate questions from text. For instance, \citeauthor{RecentQG}~\shortcite{RecentQG} mainly review traditional Seq2Seq QG models predating 2019, while \citeauthor{TOIS21}~\shortcite{TOIS21} offer an expansive overview of both traditional and pre-trained Seq2Seq models for question generation from text. 
To the best of our knowledge, our survey is the first to provide a comprehensive review of NQG across various input modalities, including knowledge bases, texts and images.

Moving forward, we first outline essential background settings for NQG in Section~\ref{sec:background}. We then introduce a new ontology for NQG, delving into structured, unstructured, and hybrid NQG in Sections~\ref{sec:stru_qg}, ~\ref{sec:un_qg} and ~\ref{sec:hybrid_nqg}, respectively. 
Furthermore, we present a thorough prospect on several promising research directions for future studies on NQG in Section~\ref{sec:fd}.

\section{Background}
\label{sec:background}
\subsection{Problem Formulation}
The NQG task aims to automatically generate textual questions from diverse input modalities, such as knowledge bases, texts, and images, which we denote as $\mathcal{X}$. 
Given an input $\mathcal{X}$, and optionally a specific target answer $A$, the objective of the NQG task is to learn a mapping function $f_\theta$ to generate a textual question $Q$. 
This is achieved by optimizing the model parameter $\theta$ to maximize the conditional likelihood $P_{\theta}\left(Q| \mathcal{X}, A\right)$. Formally, the NQG task can be described as: 
\begin{eqnarray}
    f_{\theta}: \left(\mathcal{X}, A\right) \xrightarrow[]{} Q,
\end{eqnarray}

\noindent where $f_\theta$ generates a textual question $Q = <q_1, q_2, ..., q_n>$ comprising of  a sequence of word tokens $q_i$. 
Each token $q_i$ is selected from a predefined vocabulary  $\mathcal{V}$. 
In this survey, the model $f_\theta$ is realized in various neural network architectures, encompassing Recurrent Neural Networks (RNN), Transformer, PLMs, or even LLMs.

\begin{table*}[ht]
\small
    \centering
    \renewcommand{\arraystretch}{0.8}
       \begin{tabular}{clcccc}
        \toprule
        Problem  & \multicolumn{1}{c}{Dataset} & Question Types& \#Questions& \#Ent. / \#Doc. / \#Ima. & \#Avg.Tri. / \#Avg.Que.\\
        \midrule
       \multirow{3}{*}{\centering KBQG} & WQ~\cite{MHQG}& Multi-hop& 22,989  & 25,703& 5.8\\
        &  PQ~\cite{PQ}&	Multi-hop&	9,731&	7,250&2.7\\
        & GQ~\cite{GrailQA}&	Multi-hop&	64,331&	32,585& 1.4\\
        \midrule
       \multirow{5}{*}{\centering TQG}& SQuAD~\cite{SQuAD}&	Factoid	&	97,888&20,958&4.67\\
        &MS MARCO~\cite{MARCO}&	Factoid&	3,563,535 & 1,010,916&	3.53\\
        &NewsQA~\cite{NewsQA}&	Factoid&	119,633& 12,744& 9.39\\
        &HotpotQA~\cite{HotpotQA}&	Multi-hop&	112,779& 5,000&	8\\
        &CoQA~\cite{CoQA}&	Conversational&	127,000& 8,000&	10\\
        \midrule
        \multirow{2}{*}{\centering VQG}&VQA~\cite{VQA2}&Factoid	&	369,861& 204,721& 3\\
        &VQG~\cite{VQGCOCO}&	Commonsense	&	25,000&5,000& 5\\
       \bottomrule
    \end{tabular}
    \caption{Summary of popular datasets for neural question generation. \#Questions represents the total number of questions. \#Ent., \#Doc., and \#Ima. denote the total number of entities, documents, and images in KBQG, TQG, and VQG, respectively. \#Avg.Tri. is the average number of triples in each question for KBQG. \#Avg.Que. denotes the average number of questions per document in TQG or per image in VQG.}
    \label{tab:dataset}
\end{table*}

\subsection{Popular Datasets}
We summarize popular datasets in NQG tasks, as shown in Table~\ref{tab:dataset}.
Since QG can be viewed as a dual task of QA, QG datasets are typically derived from QA datasets. 
To enhance clarity, we classify these datasets based on their input types, encompassing those derived from knowledge bases, natural language text, and visual sources.

\vpara{Knowledge Base-based Datasets.}We show three popular datasets for knowledge base question generation (KBQG).
Firstly, WebQuestions (WQ)~\cite{MHQG} comprises 22,989 instances from WebQuestionsSP~\cite{WQ} and ComplexWebQuestions~\cite{CWQ}, both of which are benchmarks for knowledge base question answering (KBQA).
These benchmarks contain questions, answers, and corresponding SPARQL queries. 
Following ~\cite{MHQG}, they convert a SPARQL query to a subgraph. 
Consequently, each instance in the WQ dataset includes subgraphs, answers, and questions. 
In addition, PathQuestions (PQ)~\cite{PQ} is constructed using two subsets of Freebase.
Notably, in PQ, the KB subgraph forms a path between the topic entities and answer entities, typically spanning connections of 2-hop or 3-hop.
Furthermore, GrailQA (GQ)~\cite{GrailQA} is a large-scale, high-quality KBQA dataset. 
Each question is associated with an S-expression, which can be interpreted as a logical form.

\vpara{Text-based Datasets.} We introduce five classical benchmark datasets for text-based question generation (TQG).
Firstly, Stanford Question Answering Dataset 
 (SQuAD)~\cite{SQuAD} is a typical reading comprehension dataset, consisting of QA pairs.
These pairs are created by crowd workers using  Wikipedia articles.
The answers to these questions are text segments extracted from the corresponding reading passages within these articles.
Secondly, MicroSoft MAchine Reading COmprehension (MS MARCO)~\cite{MARCO} is a comprehensive real-world dataset for reading comprehension.
Each question receives a response from a crowdsourced worker, ensuring that each answer is human-generated.
Thirdly, NewsQA~\cite{NewsQA} is a challenging machine comprehension dataset. Crowd workers provide questions and their corresponding answers based on news articles. The answers are composed of specific text spans extracted directly from the related news articles.
Fourthly, HotpotQA~\cite{HotpotQA}, a multi-hop QA dataset, consists of QA pairs sourced from Wikipedia. To create these pairs, crowd workers are presented with a variety of contextual supporting documents. They are instructed to formulate questions that require reasoning across these documents. Following this, they answer the questions by identifying and extracting pertinent text spans from the given context.
Lastly, CoQA~\cite{CoQA}, a large-scale conversational QA dataset, contains 127,000 QA pairs derived from 8,000 conversations. These pairs are based on text passages spanning seven diverse domains.

\vpara{Visual-based Datasets.} We present two widely used datasets for visual question generation (VQG).
Initially, VQA~\cite{VQA2}, a classical VQG benchmark, consists of images along with corresponding questions and answers. Notably, due to the unavailability of answers for the VQA test set, the validation set is commonly utilized as a proxy for test set evaluation. 
Subsequently, VQG COCO~\cite{VQGCOCO} showcases naturally formulated and engaging questions that are based on common sense reasoning. These human-annotated questions originate from the Microsoft common objects in context dataset.

\subsection{Evaluation}
Given the inherent complexity and diversity in human evaluation, we mainly focus on automatic evaluation for NQG.
We present automatic metrics across three categories, including n-grams-based, diversity, and semantic similarity metrics.

\vpara{N-gram-based Metrics.} We showcase three classical evaluation metrics that assess the n-gram similarity between the ground-truth and the generated questions.
\begin{itemize}[leftmargin=*]
\setlength{\itemsep}{0pt}
  \setlength{\parskip}{0pt}
    \item \textbf{BLEU.} BiLingual Evaluation Understudy (BLEU)~\cite{BLEU} metric evaluates the average n-gram precision against the reference text, applying a penalty for excessively short text. BLEU-$n$ calculates the proportion of the common n-grams between the generated question and the ground-truth.
    \item \textbf{ROUGE.} Recall-Oriented Understudy for Gisting Evaluation (ROUGE)~\cite{ROUGE}  focuses on recall, measuring the ratio of n-grams from the ground-truth question that are also present in the generated question.

    \item \textbf{METEOR.} Metric for Evaluation of Translation with Explicit ORdering (METEOR)~\cite{METEOR}  offers a more comprehensive assessment than BLEU. METEOR is calculated based on the harmonic mean of the unigram precision and recall, providing a balanced evaluation.
    
\end{itemize}

\vpara{Diversity Metrics.} The diversity metric is essential for tasks involving diversified question generation. A widely recognized metric, \textbf{Distinct-$n$}~\cite{Distinct}, calculates the proportion of unique n-grams in the generated question.

\vpara{Semantic Similarity Metrics.} Beyond word-level comparison, it is vital to assess sentence-level comparison, particularly about semantic similarity.
\textbf{BERTScore}~\cite{BERTScore}, a widely-used metric, utilizes pre-trained contextual embeddings from BERT to compare words between generated and ground-truth questions, calculating their similarity using cosine similarity.

\subsection{Applications}
NQG tasks have emerged as a versatile tool in various applications, each demonstrating its unique value and potential. We examine four key applications:

\vpara{Question Answering.} This application revolves around deriving answers from data sources like Wikipedia and knowledge bases in response to natural language questions. The effectiveness of QA models is largely dependent on the richness of available QA pairs. Manual labeling, a standard method for creating QA datasets, is both resource-intensive and time-consuming, which often limits the size of the datasets. QG, serving as a dual task to QA, significantly enhances the capabilities of QA systems by producing vital training data. 
For example, ~\citeauthor{DSM}~\shortcite{DSM} substitute the original questions in the WebQuestionSP dataset with questions generated by the proposed QG model, demonstrating that the questions produced by the QG model are quite close to the original questions.
In practical terms, this means more efficient information retrieval from digital assistants and more robust responses in customer service chatbots.

\vpara{Intelligent Tutoring.} Personalized education is a rapidly growing field, and intelligent tutoring systems stand at its forefront. The ability to generate custom questions tailored to a student's learning material and progress is invaluable. QG facilitates this by creating diverse and level-appropriate questions, thereby offering a more dynamic and responsive learning experience. For instance, the Bull2Sum system, developed by ~\citeauthor{QAGE}~\shortcite{QAGE} not only generates relevant questions but also contributes to a substantial educational dataset.  
In practical application, this means students receive more engaging, varied, and effective learning tools, leading to better understanding and retention of material.

\vpara{Conversational Systems.} In the realm of interactive technology, conversational systems such as virtual assistants and customer support chatbots rely heavily on QG to maintain engaging and relevant dialogues. QG enhances these systems' ability to ask contextually related and engaging questions, thereby significantly improving user interactions. The concept of conversational question generation (CQG) introduced by \citeauthor{ConQG}\shortcite{ConQG} underscores the importance of context-aware and history-informed questioning in making these interactions more natural and smooth. 

\vpara{Fact Verification.} In an era of information overload, fact verification is essential, particularly in journalism, legal investigation, and content moderation on social media. The ability of QG to generate (evidence, claim) pairs marks a significant advancement, as it automates the creation of training data for fact-checking models. The approach presented by \citeauthor{ZSFVC}\shortcite{ZSFVC} showcases how QG can streamline the validation of claims against available evidence, thereby enhancing the efficiency and accuracy of fact-checking operations.

\section{Structured Neural Question Generation}
\label{sec:stru_qg}
Structured NQG is designed to create pertinent questions based on structured data sources. Within these, the knowledge base stands out as the most typical data source. 
This section primarily focuses on \textit{knowledge base question generation} (KBQG).
KBQG generates questions based on a set of facts from a KB subgraph, where each fact is usually represented as a triple.
As shown in Figure~\ref{fig:taxonomy}, we classify KBQG models into three categories based on their architectural design: \textit{Traditional Seq2Seq models}, \textit{Graph2Seq models}, and \textit{Pre-trained Seq2Seq models}.

\subsubsection{Traditional Seq2Seq Models}
Previous studies predominantly follow the sequence-to-sequence (Seq2Seq) framework, wherein the linearized subgraph is initially inputted into an encoder to derive its representation, followed by employing a decoder to produce the question from this representation.
We categorize Seq2Seq models into two types based on the 
encoder and decoder used, namely \textbf{RNN-based} and \textbf{Transformer-based}.

\vpara{RNN-based.}  ~\citeauthor{FirstRNN}~\shortcite{FirstRNN} first use a recurrent neural network (RNN) with an attention mechanism to map the KB facts into corresponding natural language questions.
Subsequently, ~\citeauthor{K2Q-RNN}~\shortcite{K2Q-RNN} develop an RNN-based approach, K2Q-RNN, for generating questions from a specific set of keywords. Specifically, K2Q-RNN first utilizes a question keywords extractor to derive a set of keywords from entities in KB. It then employs an RNN encoder to transform these keywords into a representation. Finally, it decodes this representation to produce the output question sequence.
To improve the generalization for KBQG, ~\citeauthor{Zero-ShotQG}~\shortcite{Zero-ShotQG} propose an encoder-decoder framework leveraging extra textual contexts of triples. Concretely, a feed forward architecture encodes the input triple and a set of  RNN to encode textual context. A  decoder is equipped with triple and textual context attention modules and a copy mechanism to generate questions.
Despite these advances, two classical challenges remain: limited information and semantic drift. To solve these, ~\citeauthor{KTG}~\shortcite{KTG} design a novel model, KTG, which integrates a knowledge-augmented fact encoder and a typed decoder within a reinforcement learning framework, enhanced by a grammar-guided evaluator.
The encoder processes entities, relations, and auxiliary knowledge to create an augmented fact representation used by the decoder for question generation, while the evaluator assesses each question's grammatical similarity to the ground-truth, providing feedback that continually refines the encoder-decoder module.

\vpara{Transformer-based.}
Considering the limitations of RNN in terms of effectiveness and efficiency in handling larger contexts, Transformer~\cite{transformer} emerges as a robust and effective alternative, offering a promising solution.
~\citeauthor{MHQG}~\shortcite{MHQG} propose a novel model to generate complex multi-hop, difficulty-controllable questions from subgraphs.
The encoder encodes the difficulty level of the given subgraph, subsequently enabling the decoder to generate questions that are tailored to this specific difficulty level.
To effectively generate questions that not only articulate the specified predicate but also correlate with a definitive answer, ~\citeauthor{IDCQG}~\shortcite{IDCQG} utilize a range of diverse contexts and design an answer-aware loss.
A context-augmented fact encoder, equipped with multi-level copy mechanisms, effectively captures the diversified information across contexts and fact triples, thereby mitigating the issue of inaccurate predicate expression.
Furthermore, the use of an answer-aware loss function ensures that the generated questions align with definitive answers by applying cross-entropy between the words in the question and those that denote the answer type.

\subsubsection{Graph2Seq Models}
The intricate structural information within a KB is crucial for generating high-quality questions in KBQG. 
However, previous approaches fall short in effectively capturing this rich structural information, as they merely linearize a KB subgraph into a sequence of triples and use RNN / Transformer models to learn its embeddings.
Motivated by this, ~\citeauthor{G2S}~\shortcite{G2S} propose a novel Graph-to-Sequence (Graph2Seq) model, which first utilizes a bidirectional gated graph neural network-based (BiGGNN-based) encoder to encode the KB subgraph, followed by decoding the output question using an RNN-based decoder, equipped with a node-level copy mechanism.
~\citeauthor{Meta-CQG}~\shortcite{Meta-CQG} propose a meta-learning framework, Meta-CQG, to address the data imbalance issue.
They initially utilize graph-level contrastive learning to train a graph retriever, followed by retrieving similar subgraphs using cosine similarity across graph embeddings. Subsequently, they employ a meta-learning approach to train a generator tailored to each input subgraph by learning the potential features of retrieved similar subgraphs.

\subsubsection{Pre-trained Seq2Seq Models}
Pre-trained language models (PLMs), pre-trained on a large-scale corpus, possess rich semantic knowledge, which can boost the performance of downstream KBQG tasks through fine-tuning.
With the continuous expansion in parameter size and training corpus volume, the field has witnessed the evolution of large language models (LLMs).
In light of this, we categorize pre-trained Seq2Seq models for KBQG into two groups: those based on traditional PLMs (\textbf{PLMs-based}) and those utilizing the more advanced LLMs (\textbf{LLMs-based}).

\vpara{PLMs-based.} 
BART~\cite{BART} and T5~\cite{T5}, two widely recognized PLMs, are built upon the encoder-decoder framework. 
A significant challenge in adapting these PLMs to KBQG tasks lies in bridging the semantic gap. This is primarily because PLMs are originally pre-trained on unstructured text, which contrasts with the structured nature of KB. Additionally, another key challenge involves effectively capturing the structural information inherent in KB.

To address the above challenges, ~\citeauthor{JointGT}~\shortcite{JointGT} introduce three pre-training tasks: graph-enhanced text reconstruction, text-enhanced graph reconstruction, and graph-text embedding alignment to explicitly build the connection between knowledge graphs and text sequences.
Additionally, they create a semantic aggregation module that is aware of the structure at every Transformer layer, which gathers contextual information according to the structure of the graph.
~\citeauthor{DSM}~\shortcite{DSM} address the issue of the semantic gap by converting the input subgraph into a linear sequence of triples, achieved through the concatenation of relational triples. Furthermore,  they explicitly incorporate structural information as input. This incorporation involves directly inserting special tokens “〈H〉”, “〈R〉”, and “〈T〉” before the head entity, relation, and tail entity in each fact triple, respectively. This approach effectively clarifies the relationships between entities, ensuring a more coherent representation within the model.
Likewise, ~\citeauthor{DiversifyQG}~\shortcite{DiversifyQG} add special tokens indicative of structured information at the beginning of each element within every triple in KB.
~\citeauthor{AutoQGS}~\shortcite{AutoQGS} focus on directly generating questions from SPARQL, aimed at covering complex operations.
They first execute the SPARQL query on a KB to retrieve a corresponding subgraph. This subgraph is then linearized, serving as input for an auto-prompter, which generates the prompt text. Subsequently, this prompt text, along with the original SPARQL query, are used as inputs for a QG generator. This process ultimately results in high-quality question generation, effectively bridging the gap between non-natural language SPARQL and the natural language question.

\vpara{LLMs-based.}
Despite the success of PLMs-based models on KBQG~\cite{LFKQG,DSM,DiversifyQG}, their effectiveness hinges on extensive fine-tuning using large training datasets.
However, the creation of labeled datasets is costly and time-consuming. Hence, researchers are increasingly focusing on few-shot KBQG tasks to mitigate these challenges~\cite{AutoQGS}.
Recently, LLMs, like ChatGPT and Llama2, have exhibited exceptional capabilities in various few-shot and zero-shot tasks.
This emerging insight inspires researchers to investigate few-shot KBQG tasks, leveraging the capabilities of LLMs.
The key challenge is to design effective prompts that prompt LLMs to generate the targeted questions for KBQG.
~\citeauthor{KQG-COT}~\shortcite{KQG-COT} propose KQG-COT framework, which involves first employing LLMs (i.e., text-davinci-003) to generate ideal questions for KBQG using chain-of-thought (COT).
To be specific, KQG-COT first identifies suitable logical forms from the unlabeled data pool, meticulously evaluating their attributes. Following this, a specialized prompt is developed to steer LLMs in creating complex questions derived from these chosen logical forms.
~\citeauthor{SGSH}~\shortcite{SGSH} develop a fine-grained prompting approach named SGSH. Concretely, SGSH involves training a learnable skeleton generator that then uses the generated skeleton to create skeleton-based prompts, effectively stimulating LLMs to generate desired questions.

\section{Unstructured Neural Question Generation}
\label{sec:un_qg}
Unstructured NQG focuses on producing textual questions derived from unstructured data sources such as texts and visual images. Accordingly, our exploration will specifically focus on two distinct types: Text-based Question Generation (\textbf{TQG}) and Visual Question Generation (\textbf{VQG}).

\subsection{TQG}
As shown in Figure~\ref{fig:taxonomy}, TQG models are primarily divided into three types: \textit{Traditional Seq2Seq models}, \textit{Graph-based Models}, and \textit{Pre-trained Seq2Seq models}.

\subsubsection{Traditional Seq2Seq Models}
Most TQG models adhere to the Seq2Seq framework. This framework first employs an encoder to compress the input text into low-dimensional vectors that retain the essential semantic meanings. Subsequently, a decoder is employed to generate questions based on these condensed vectors.
We divide TQG models into RNN-based and Transformer-based models according to their backbone architecture.

\vpara{RNN-based.}\citeauthor{NQGRC}~\shortcite{NQGRC} first apply RNN to TQG tasks, leveraging an attention mechanism to enable the decoder to concentrate on the most pertinent segments of the input text.
In the process of determining the specific information to emphasize while generating questions, the majority of Seq2Seq models utilize the features of answer positions to integrate the spans of answers.
For example, ~\citeauthor{HybridNQG}~\shortcite{HybridNQG} contend that context words near the answer are more apt to be answer-relevant. Hence, they explicitly encode the positional proximity of these context words to the answer by position embedding and a position-aware attention mechanism.
Nevertheless, ~\citeauthor{QGWPC}~\shortcite{QGWPC} think the proximity-based approach does not always work. Therefore,
they devise a more generalized model, which exploits answer-relevant relations to facilitate the faithfulness of the generated question.

However, when dealing with long documents as the input context, these models face increased difficulty in effectively exploiting relevant content while avoiding irrelevant information.
To solve this issue, ~\citeauthor{CorefNQG}~\shortcite{CorefNQG} suggest integrating coreference knowledge into the encoder to improve the model's ability to identify entities across different sentences, thereby enhancing the quality of question generation.
~\citeauthor{CGCQG}~\shortcite{CGCQG} apply multi-stage attention to focus on crucial segments of the document that are pertinent to the answer, leveraging them to facilitate the generation of questions.

\vpara{Transformer-based.} Due to the inherent sequential nature of RNN, RNN-based models face significant computational costs and struggle with long-range dependency issues. Fortunately, the Transformer effectively addresses these challenges, resulting in the widespread adoption of Transformer-based models for TQG tasks.
~\citeauthor{CLQG}~\shortcite{CLQG} develop a cross-lingual model designed to enhance QG for a primary language by utilizing resources from a secondary language. 
~\citeauthor{NQGAP}~\shortcite{NQGAP} regard the answer as the hidden pivot for QG. Specifically, they first generate the hidden answer according to the paragraph. Subsequently, they merge this paragraph with the derived pivot answers to generate the question.
~\citeauthor{SQG}~\shortcite{SQG} present a semi-autoregressive approach for generating sequential questions. Concretely, they segment the target questions into various groups, and then simultaneously generate each group of closely related questions.

\subsubsection{Graph-based Models} Traditional Seq2Seq models struggle to capture the inherent structure of context, including syntax and semantic relationships. In contrast, graph neural networks (GNNs), with their inherent advantage in graph structure, are more adept at understanding and expressing the relationships between entities or sentences.
Consequently, researchers are increasingly adopting GNN-based models for TQG tasks~\cite{G2S-RL,MulQG,SGGDQ,IGND,CQG}.
In general, most approaches initiate by constructing a graph from the input context, followed by utilizing a GNN to learn the graph representation from the constructed text graph effectively. Subsequently, this representation is fed into the decoder to produce the question.
For instance, ~\citeauthor{G2S-RL}~\shortcite{G2S-RL} first create two types of passage graph from the input text, \emph{i.e.}, syntax-based static graph and semantics-aware dynamic graph.
Following this, they introduce an innovative bidirectional gated graph neural network, designed to effectively learn the passage graph embeddings from the assembled text graph.
~\citeauthor{SGGDQ}~\shortcite{SGGDQ} focus on generating deep questions, wherein they initially extract key information from the passage to organize it as a semantic graph. Subsequently, they propose an attention-based gated graph neural network to capture the dependency relations of the semantic graph.
~\citeauthor{IGND}~\shortcite{IGND} propose a novel model, in which a relational-graph encoder is introduced for encoding dependency relations within passages, accompanied by an iterative GNN-based decoder. This decoder is specifically designed to capture structural information throughout each step of the generation process.
~\citeauthor{CQG}~\shortcite{CQG} first construct an entity graph from the input documents, followed by utilizing a graph attention network to extract key entities. Furthermore, they introduce a controlled Transformer-based decoder, enhanced with a flag tag, to ensure the inclusion of these key entities in the generated questions.

\subsubsection{Pre-trained Seq2Seq Models}
Pre-trained language models, through pre-training on vast textual corpora, acquire an extensive range of linguistic knowledge, which can significantly enhance the performance of downstream tasks. 
While PLMs exhibit remarkable proficiency in processing natural language text, there still exist several challenges for TQG tasks.
Primarily, PLMs are not specifically trained on TQG datasets, leading to their reduced proficiency in TQG tasks.
Additionally, PLMs rely on generic self-supervised learning tasks, which are not tailored for TQG tasks, resulting in suboptimal performance in TQG tasks.

To address these challenges, researchers are increasingly focusing on fine-tuning PLMs for specific downstream tasks and adapting their neural architectures accordingly~\cite{UNILM,MiniLM,ERNIE-GEN,CoHS-CQG,MultiFactor}. 
For example, ~\citeauthor{UNILM}~\shortcite{UNILM} present a unified pre-trained language model (UNILM), which is distinctively optimized across three distinct types of language modeling tasks, including unidirectional, bidirectional, and sequence-to-sequence prediction. 
~\citeauthor{BertGen}~\shortcite{BertGen} propose a novel pre-training approach tailored specifically for QG tasks. This approach intensively focuses on the answer, aiming to generate contextually relevant sentences containing missing answers. By doing this, it aims to learn more effective representations that are highly optimized for the question generation task.
~\citeauthor{SG-CQG}~\shortcite{SG-CQG} present a new framework comprising two modules for generating conversational questions: ``what-to-ask" and ``how-to-ask".  The ``what-to-ask" module constructs a semantic graph to extract underlying rationale and selects the relevant answer span. The ``how-to-ask" module uses a classifier to identify the appropriate question type. Subsequently, the framework fine-tunes the T5~\cite{T5} model on the tailored dataset to produce conversational questions.
~\citeauthor{MultiFactor}~\shortcite{MultiFactor} introduce phrase-enhanced Transformer, an effective model that capitalizes on the strengths of powerful PLMs.
This method creatively integrates phrase selection probabilities from the encoder into the decoder, significantly enhancing the quality of question generation.

\subsection{VQG}
VQG may be considered a dual task of visual question answering.
As illustrated in Figure~\ref{fig:taxonomy}, VQG models are typically divided into three main categories, including \textbf{CNN-based}, \textbf{GNN-based}, and \textbf{Generative Networks-based}.

\vpara{CNN-based.} Most VQG models typically employ a convolutional neural network (CNN) to encode an image and a RNN to encode an answer, both merging into an intermediate representation. This representation is then decoded to generate a question.
For example, ~\citeauthor{iVQA}~\shortcite{iVQA} devise a multi-model attention module to dynamically identify regions in the image that are relevant to the answer.
To generate difficulty-controllable questions, ~\citeauthor{DGN}~\shortcite{DGN} introduce a difficulty control mechanism in the decoder, utilizing a difficulty variable to regulate the complexity of the questions generated.
~\citeauthor{KECVQG}~\shortcite{KECVQG} present a knowledge-enhanced causal visual question generation (KECVQG) model, which addresses the inherent bias in previous VQG models by employing a causal approach and knowledge integration to generate more accurate and unbiased questions from images.

\vpara{GNN-based.} One key challenge of VQG is to focus on answer-related regions during question generation.
To solve the issue, researchers propose several approaches to perform explicit region selection.
For instance, ~\citeauthor{Radial-GCN}~\shortcite{Radial-GCN} perform explicit object-level cross-modal interaction
by identifying a core answer area and constructing an answer-related graph convolutional network (GCN) graph structure.
~\citeauthor{MOAG}~\shortcite{MOAG} leverage a co-attention network and a graph network to identify and relate key objects in an image to a target answer, thereby generating more comprehensive questions.
However, previous approaches rely solely on semantic features to identify regions related to the answer, leading to potential biases and overlooking complex relations between objects.
Given this, ~\citeauthor{MCMI}~\shortcite{MCMI} utilize contrastive learning to integrate semantic knowledge with regional representations and leverage a relation-level interaction scenario to consider various types of relationships between regions and answers.

\vpara{Generative Networks-based.} To overcome the limitation of existing VQG models to produce generic and uninformative questions, ~\citeauthor{IMVQG}~\shortcite{IMVQG} introduce a novel method that maximizes the mutual information between the image, the expected answer, and the generated question. This is achieved by employing a Variational Auto-Encoder (VAE) framework and utilizing two distinct latent spaces, enhancing the diversity and relevance of the questions generated.
Observing that previous VQG models often rely heavily on answers, leading to overfitting and a lack of creativity. ~\citeauthor{C3VQG}~\shortcite{C3VQG} propose a category-specific, cyclic training approach. This innovative method employs weak supervision and structured latent spaces, enabling the generation of diverse and relevant questions based on categories, thereby eliminating the need for ground-truth answers.
\section{Hybrid Neural Question Generation}
\label{sec:hybrid_nqg}
Hybrid NQG aims to generate textual questions based on both structured and unstructured data sources~\cite{wang2023multiqg,ConVQG,Unify_FQG}, where multimodal is common. 
Compared with the previous single-modal question generation, hybrid question generation is more prevalent in real-life scenarios, especially in the education field.
The primary challenge in hybrid question generation lies in effectively integrating information across diverse data sources or modalities.
As pioneers in the field, \citeauthor{wang2023multiqg}~\shortcite{wang2023multiqg} first investigate multi-modal question generation from images and texts, proposing a novel and effective PLMs-based approach that surpasses the performance of ChatGPT.
\citeauthor{Unify_FQG}~\shortcite{Unify_FQG} propose a unified framework for generating contextual questions (CQG) and factoid questions (FQG), which addresses current methods' limitations in structural and contextual information. 
Specifically, they introduce shared task modules for cross-domain learning and task-specific modules that integrate external knowledge for CQG and enhance contextual understanding for FQG, demonstrating advanced performance. Despite these promising results, hybrid NQG remains under-explored, with significant potential for further innovation and improvement.

\section{Conclusion and Future Directions}
\label{sec:fd}
This paper provides a thorough overview of Neural Question Generation (NQG) in various modalities. We first introduce popular datasets, classical evaluation metrics, and four prominent applications. We then explore prevalent methods for modeling diverse inputs, including structured NQG, unstructured NQG, and hybrid NQG.
Despite the notable achievements of NQG models, several challenges remain, suggesting promising directions for future research.

\vpara{Proactive Question Generation.} Previous studies predominantly focus on producing reactive questions based on the provided inputs.
However, the ability to proactively tailor question generation to meet specific user requirements and achieve pre-defined targets is essential in real-world applications.
Intelligent tutoring systems serve as a prime example, engaging students with tailored interactions. 
These systems carefully design a series of exercises aimed at specific objectives, gradually guiding step by step from simple to advanced towards the targeted goal, in order to enhance students' understanding of specific concepts.
Although the benefits and practical applications are evident, the field of proactive question generation remains under-explored. This underlines its significant potential as a promising field for future research.

\vpara{Multi-modal Question Generation.} Current question generation tasks primarily concentrate on single-modal question generation, such as KBQG, TQG, and VQG. Nevertheless, in practical scenarios, the significance of multi-modal question generation is on the rise. This is particularly evident in the educational field, where numerous scientific questions require an understanding of both visual images and textual descriptions.
To the best of our knowledge, multi-modal question generation remains in a very early stage, with few studies having been conducted in this field~\cite{wang2023multiqg}, as detailed in Section~\ref{sec:hybrid_nqg}.
Given the advanced capabilities of vision-language pre-trained models like CLIP, a compelling research direction is to develop effective strategies for leveraging VL-PLMs in multi-modal question generation.

\vpara{Controllable Question Generation.} The capacity to control specific aspects of question generation holds significant applications, such as creating different difficulty levels~\cite{MHQG} and types of questions for intelligent tutoring systems (ITS).  This potential paves the way for future research, particularly in exploring how these customized elements can enhance personalized learning experiences within ITS.
Additionally, certain studies overlook subjective human factors such as sentiment and style, focusing instead on the influence of the input and the target answer. Yet, these human elements are critical in influencing the process of question generation. Thus, upcoming studies need to investigate approaches that correspond to distinct human behaviors and preferences.

\vpara{Automatic Evaluation Metrics for Generation.} Widely used metrics such as BLEU and ROUGE assess question quality by measuring the lexical overlap between the generated question and the ground-truth.
Yet, these metrics can potentially penalize well-formed questions that diverge in lexical similarity from the ground-truth questions, indicating a limitation in capturing question validity.
Accordingly, a more reasonable metric for assessing question quality would consider key factors such as question answerability~\cite{RQUGE}, consistency with the context provided, and containing a sufficient amount of information content.
Meanwhile, the advancement of diversity metrics is critical, particularly due to the significant diversity capabilities demonstrated by LLMs.
The popular diversity metric, Distinct-n, emphasizes the ratio of unique n-grams but it is overly simplistic.
Hence, a comprehensive assessment of diversity can be conducted from multiple perspectives, including semantic diversity~\cite{DiversifyQG}, syntactic diversity, and thematic diversity.
This highlights the need for innovative metrics to accurately evaluate the diverse aspects of question quality.

\section*{Acknowledgments}
This work is supported by the National Key Research \& Develop Plan (2023YFF0725100) and the National Natural Science Foundation of China (62322214, U23A20299, 62076245, 62072460, 62172424, 62276270). This work is supported by Public Computing Cloud, Renmin University of China.  We gratefully acknowledge the support provided by the China Scholarship Council Scholarship Fund. We sincerely thank all reviewers for their valuable feedback.


 \small
\bibliographystyle{named}
\bibliography{ijcai24}
\end{document}